\documentclass[runningheads]{llncs}
\usepackage[T1]{fontenc}

\usepackage{graphicx}
\usepackage{amsmath}
\usepackage{hyperref}
\usepackage{color}
\usepackage{booktabs}
\usepackage[skip=1.5pt]{caption}
\begin{document}
\title{SegDT: A Diffusion Transformer-Based Segmentation Model for Medical Imaging\thanks{The supports of TotalEnergies and Sorbonne University Abu Dhabi are fully acknowledged (Abdenour Hadid).}}
\titlerunning{SegDT: A Diffusion Transformer-Based Segmentation Model}

\author{Salah Eddine Bekhouche\inst{1}\orcidID{0000-0001-5538-7407} \and
Gaby Maroun\inst{1}\orcidID{0009-0006-5608-1817} \and
Fadi Dornaika\inst{1,2}\orcidID{0000-0001-6581-9680} \and Abdenour Hadid\inst{3}\orcidID{0000-0001-9092-735X}}

\authorrunning{SE. Bekhouche et al.}
%
\institute{University of the Basque Country UPV/EHU, San Sebastian, Spain \\
\email{\{sbekhouche001,gmaroun001\}@ikasle.ehu.eus} \email{fadi.dornaika@ehu.eus} \and
IKERBASQUE, Basque Foundation for Science, Bilbao, Spain \and
Sorbonne University Abu Dhabi, Abu Dhabi, UAE \\
\email{abdenour.hadid@sorbonne.ae}
}
\maketitle              
\begin{abstract}
Medical image segmentation is crucial for many healthcare tasks, including disease diagnosis and treatment planning. 
One key area is the segmentation of skin lesions, which is vital for diagnosing skin cancer and monitoring patients. 
In this context, this paper introduces SegDT, a new segmentation model based on diffusion transformer (DiT). SegDT is designed to work on low-cost hardware and incorporates Rectified Flow, which improves the generation quality at reduced inference steps and maintains the flexibility of standard diffusion models. Our method is evaluated on three benchmarking datasets and compared against several existing works, achieving state-of-the-art results while maintaining fast inference speeds. This makes the proposed model appealing  for real-world medical applications. This work advances the performance and capabilities of deep learning models in medical image analysis, enabling faster, more accurate diagnostic tools for healthcare professionals. The code is made publicly available at \href{https://github.com/Bekhouche/SegDT}{GitHub}.

\end{abstract}

\vspace{-2mm}
\section{Introduction}
\label{sec:introduction}
\vspace{-2mm}

Skin cancer is a major global health concern, and its early detection is crucial for improving survival rates. Medical image segmentation plays a vital role in this process by enabling precise lesion boundary delineation. Among the successful approaches to automatic image segmentation is deep learning which has revolutionized the field, with Convolutional Neural Networks (CNNs) like U-Net \cite{ronneberger2015unet} and DeepLabV3+ \cite{chen2018deeplab} demonstrating interesting performance. However, CNNs can struggle with long-range dependencies, limiting their ability to segment complex or irregularly shaped lesions. As an alternative, Transformers \cite{vaswani2017attention}, inspired by their success in Natural Language Processing (NLP) and vision, offer an appealing solution by capturing global context through self-attention, with models like TransUNet \cite{chen2021transunet} and Swin-UNet \cite{cao2022swin} showing improved accuracy (although their high computational cost can hinder real-world applications). Diffusion models, on the other hand, have recently emerged as a powerful technique, achieving state-of-the-art results in various medical imaging tasks \cite{wu2024medsegdiff} by iteratively refining segmentation through a denoising process. While highly accurate, their computational cost and long inference times pose a challenge for real-world deployment.

To address some of these limitations, we propose SegDT, an extra small Diffusion Transformer (DiT) model designed for efficient segmentation on low-cost GPUs. SegDT integrates rectified flow to accelerate inference while maintaining high segmentation accuracy. 

The main contributions of this work are: {\bf (i)} A compact DiT architecture optimized for resource-constrained GPUs is proposed; {\bf (ii)} A methodology for the incorporation of rectified flow for efficient inference with reduced sampling steps is described; and {\bf (iii)} Extensive experiments are conducted on three benchmarking datasets, achieving state-of-the-art performances, compared to exiting methods.

\vspace{-4mm}
\section{Related Work}
\label{sec:related_work}

This section discusses relevant prior work in medical image segmentation, focusing on CNNs, Transformers, hybrid architectures, and diffusion models, with an emphasis on approaches related to skin lesion segmentation and efficient architectures.

\vspace{-2mm}
\subsection{CNN-based Segmentation}
\vspace{-2mm}
CNNs have been the dominant approach in medical image segmentation for many years.  U-Net \cite{ronneberger2015unet}, with its encoder-decoder structure and skip connections, revolutionized the field by enabling effective learning with limited data.  Its success stems from the ability to capture both local and global context, crucial for accurate segmentation.  Many variants and extensions of U-Net have been proposed, such as DeepLabV3+ \cite{chen2018deeplab}, which incorporates 
convolutions to capture multi-scale information, and ResUNet++ \cite{jha2019resunet++}, which leverages residual connections for improved training stability.  However, CNNs, due to their limited receptive field, often struggle with capturing long-range dependencies, which can be critical for segmenting complex or irregularly shaped lesions, like those encountered in skin cancer imaging. While these methods can  achieve impressive results, their limitations motivated researchers to introduce Transformers based methods.

\vspace{-2mm}
\subsection{Transformer-based Segmentation}
\vspace{-2mm}

Transformers, initially developed for NLP \cite{vaswani2017attention}, have recently shown remarkable success in computer vision \cite{dosovitskiy2020image}.  Their key strength lies in the self-attention mechanism, which allows them to capture long-range dependencies and global context effectively.  In medical image segmentation, TransUNet \cite{chen2021transunet} was a pioneering work, combining the strengths of transformers and U-Nets.  It utilizes a transformer encoder to extract global features and a U-Net decoder to preserve local details.  While TransUNet demonstrated the potential of transformers, its computational complexity can be a limiting factor. Swin-UNet \cite{cao2022swin} addressed this issue by introducing a hierarchical Swin Transformer architecture with shifted windows, enabling efficient computation of self-attention.  This approach balances local and global feature extraction, achieving competitive performance with reduced computational cost. However, transformer-based models can still be resource-intensive, motivating the development of more resource- efficient architectures. 

\vspace{-2mm}
\subsection{Hybrid Architectures: CNN-Transformers}
Recognizing the complementary strengths of CNNs and transformers, researchers have explored hybrid architectures.  DS-TransUNet \cite{lin2022ds} incorporates deformable self-attention into the TransUNet framework, allowing the model to focus on relevant regions in the image. BRAU-Net++ \cite{lan2024brau} combines convolutional feature extractors with transformer-based global reasoning, achieving state-of-the-art results on several benchmarks. These hybrid approaches aim to leverage the local feature extraction capabilities of CNNs and the global context modeling of transformers.  While effective, these methods often come with increased complexity and resource requirements. MobileUNETR \cite{perera2024mobileunetr} takes a different approach by focusing on efficiency.  It proposes a lightweight transformer-based model optimized for mobile and edge devices. While sacrificing some accuracy compared to larger models, it achieves faster inference, making it suitable for resource-constrained environments.  

\vspace{-2mm}
\subsection{Diffusion Models for Segmentation}

Diffusion models have recently emerged as a powerful tool for image generation and segmentation.  They operate by progressively adding noise to an image until it becomes pure noise (forward diffusion) and then learning to reverse this process to generate the original image (reverse diffusion).  In medical image segmentation, MedSegDiff \cite{wu2024medsegdiff} integrates transformers within a diffusion framework, enabling the model to capture fine-grained anatomical details.  MedSegDiff-V2 \cite{wu2024medsegdiffv2} further improves upon this by incorporating multi-resolution features.  While these diffusion-based models have demonstrated impressive performance, their computational cost and long inference times can be a limiting factor for real-time deployment.

Our work addresses this limitation by introducing SegDT, a Diffusion Transformer (DiT) model that incorporates rectified flow to accelerate inference, making it more suitable for real-world applications. Unlike previous diffusion models that often rely on U-Net architectures, we explore a more efficient transformer-based design. Furthermore, we consider a rectified flow approach to achieve high-quality segmentations with fewer sampling steps, significantly reducing the inference time. This is a key difference between our work and previous diffusion-based segmentation methods.

\vspace{-2mm}
\section{Proposed Method}
\label{sec:proposed_method}
\vspace{-2mm}

\begin{figure}
    \centering
    \includegraphics[width=1.0\linewidth]{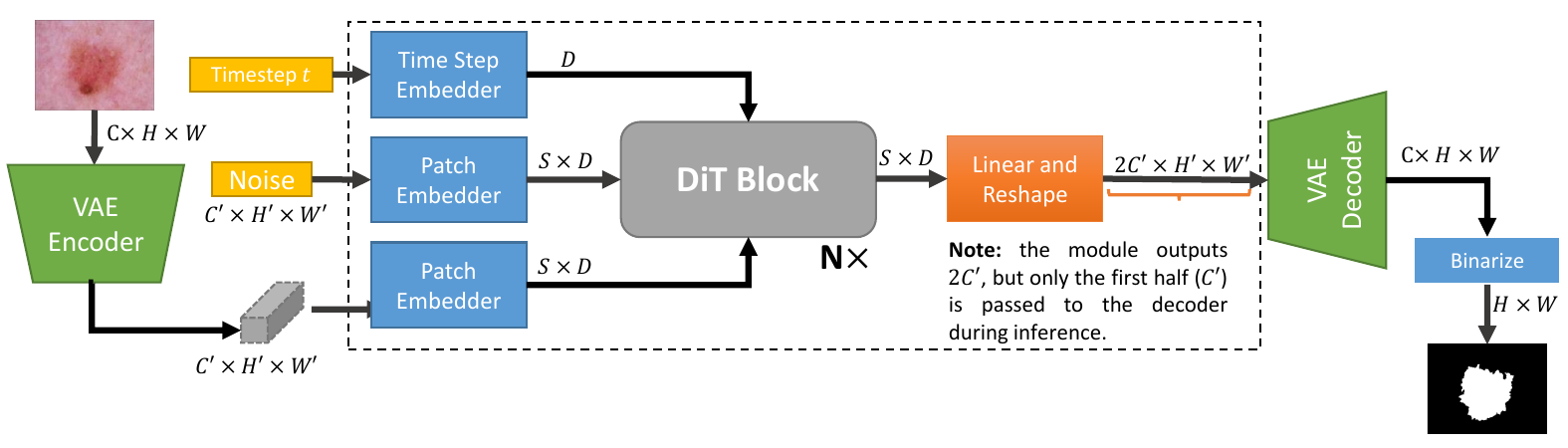}
    \caption{Overview of the SegDT inference architecture for medical image segmentation.}
    \label{fig:general_architecture}
\end{figure}

This section details SegDT, our proposed architecture for segmenting skin lesions in medical images. Fig. \ref{fig:general_architecture} illustrates the SegDT architecture, which comprises a Variational Autoencoder (VAE) encoder, a DiT, and a VAE decoder. The VAE components are based on the pretrained Tiny AutoEncoder for Stable Diffusion (TAESD)\footnote{\url{https://github.com/madebyollin/taesd}}, chosen for its compact size and computational efficiency. TAESD's small footprint and fast encoding/decoding speeds contribute to SegDT's overall efficiency by generating a compact latent representation, compressing the width and height 8 times. This smaller latent space allows for a more efficient DiT, reducing the computational burden and further accelerating the segmentation process. The DiT component is based on the DiT-XS (extra-small) variant, which employs a patch size of 2, as detailed in \cite{peebles2023scalable}. DiTs have demonstrated strong performance in image generation, making them a suitable foundation for our segmentation task. SegDT is designed to address the challenges of precise lesion boundary delineation in noisy medical images - a critical requirement for accurate dermatological diagnosis and treatment planning. Furthermore, the model's compact architecture and the integration of rectified flow during inference enable rapid and efficient analysis by reducing the required number of inference steps.

\vspace{-10mm}
\subsection{Rectified Flow for Efficient Inference}

In the forward diffusion process, a segmentation label \(x_0 \in \mathcal{X}\) (represented by its encoded latent representation \(z_0 \in \mathcal{Z}\)) is gradually perturbed by adding Gaussian noise over \(T\) timesteps. This results in a sequence of noisy segmentations \(x_1, x_2, \dots, x_T \in \mathcal{X}\) (with corresponding noisy latent representations \(z_1, z_2, \dots, z_T \in \mathcal{Z}\)). The reverse diffusion process aims to reconstruct the original segmentation mask \(x_0\) (or \(z_0\)) from the noisy \(x_T\) (or \(z_T\)), conditioned on a latent representation \(y \in \mathcal{Y}\) of the corresponding medical image.  Inspired by efficient diffusion sampling techniques like those used in Denoising Diffusion Implicit Models (DDIMs) \cite{song2020denoising}, we employ a rectified flow approach to learn a modified, more efficient reverse process.

Instead of directly predicting the noise added at each timestep, our model learns a velocity field \(v_\theta(z, t, y): \mathcal{Z} \times [0, T] \times \mathcal{Y} \rightarrow \mathcal{Z}\) in the latent space \(\mathcal{Z}\). This velocity field represents the direction and magnitude of the change required to move from \(z_t\) to \(z_{t-1}\), conditioned on \(z_t\), the time step \(t \in [0, T]\), and the latent representation of the image \(y\). The reverse diffusion process then follows this learned velocity field.  We approximate the reverse trajectory using a numerical integration method, such as Euler's method.

\begin{equation}
z_{t-1} = z_t + v_\theta(z_t, t, y) \Delta t,
\end{equation}

where \(\Delta t\) is the size of the integration step. In our normalized formulation, we assume that the continuous time interval is \([0,1]\). When this interval is divided into discrete timesteps of \(T\), each step corresponds to an increment of \(\Delta t = \frac{1}{T}.\)

The velocity field \(v_\theta\) is learned by minimizing a loss function. Given a noisy latent representation \(z_t\) (obtained by adding noise to the encoded ground truth mask \(z_0\)), the model predicts the velocity \(v_\theta(z_t, t, y)\). This prediction is compared to a target velocity derived from the known noise added to \(z_0\) to obtain \(z_t\).  A common metric for this comparison is the Mean Squared Error (MSE).  This training process enables the model to accurately predict the velocity needed to reverse the diffusion process.

Learning a velocity field, rather than directly predicting noise, can be seen as a form of rectification, promoting a smoother and more predictable reverse diffusion process.  It also provides increased flexibility for incorporating conditioning information, such as the anatomical context provided by \(y\). Although the concept of learning a velocity field for diffusion has been explored (e.g., \cite{liu2022flow}), the specific term "rectified flow" and its precise definition are still evolving within the diffusion modeling community.  Our use of this term emphasizes the goal of a more direct, efficient path from noisy latent to segmentation.

\vspace{-2mm}
\subsection{Transformer-Based Diffusion Model (SegDT)}
\vspace{-2mm}

This section describes the architecture of SegDT, a transformer-based diffusion model for image segmentation. SegDT employs a pre-trained VAE to map images into a latent space.  This latent representation is then processed by a DiT. SegDT employs 12 DiT blocks.

During training, the input to the DiT consists of latent representations of ground truth segmentation masks.  In contrast, during inference, the model generates segmentations by processing randomly sampled, noisy latent vectors initialized with a fixed seed.  Critically, the VAE decoder component is not utilized during training, as the loss function is computed directly within the latent space. This approach significantly reduces training time.

The SegDT architecture comprises the following components:

\subsubsection{VAE Encoder}
A pre-trained VAE encoder is employed to map input images of size \(C \times H \times W\) into a lower-dimensional latent space. During training, this encoder also processes corresponding ground-truth masks in the same manner. The resulting latent representation has dimensions \(C' \times H' \times W'\), where \(C'=4\), and the spatial dimensions are reduced by a factor of 8, such that \(H' = H/8\) and \(W' = W/8\). This latent representation serves as the input to the DiT model.

\subsubsection{Patch Embedder} 
A patch embedding module, utilizing a shared architecture for both input and conditional tensors, is applied. For the input, it processes the latent representation from the VAE encoder. During inference, when ground-truth masks are not available, it processes a randomized tensor for the conditional input.  The input tensor is flattened and then linearly projected to a \(S \times D\) dimensional embedding space, where \(D=192\) is the embedding dimension derived from the DiT's token length. The number of patches, \(S\), is determined by \(S = \frac{H' \times W'}{P^2}\), where \(P=2\) is the patch size used in the DiT model, and \(S=256\) represents the total number of patches.

\subsubsection{Time Step Embedder} 
A time step embedding module is used to generate a \(D\)-dimensional vector (where \(D=192\), consistent with the patch embedding dimension) that encodes the diffusion time step \(t\). This embedding vector is then used to condition the DiT blocks, providing information about the current stage of the diffusion process.

\subsubsection{DiT Blocks}
\begin{figure}
    \centering
    \includegraphics[width=1.0\linewidth, height=0.35\linewidth]{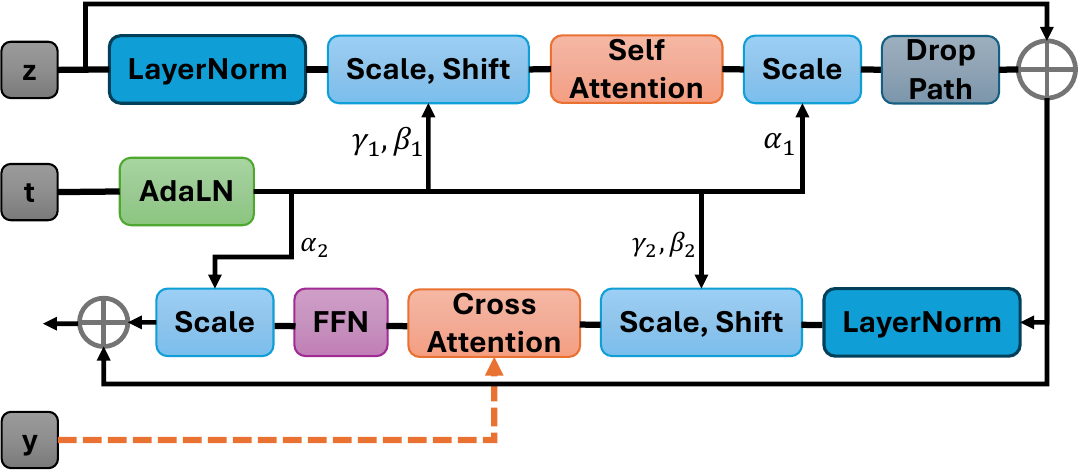}
    \caption{Detailed architecture of a single DiT block.}
    \label{fig:dit_block}
\end{figure}

A series of DiT blocks form the core of SegDT. Each block receives the embedded latent representation \(\texttt{z}\), the timestep embedding \(\texttt{t}\), and the condition embedding \(\texttt{y}\) as input.  These blocks progressively refine the latent representation \(\texttt{z}\). The detailed architecture of each DiT block is shown in Fig. \ref{fig:dit_block}. Each DiT block comprises:
\begin{itemize}
    \item \textbf{Adaptive Layer Normalization (adaLN):} Normalizes activations and then modulates them using the parameters $\alpha_1$, $\alpha_2$ (for scaling), $\beta_1$, $\beta_2$ (for shifting), and $\gamma_1$, $\gamma_2$ (for scaling) to shift and scale the normalized activations.
    \item \textbf{Self-Attention:} Models global spatial relationships within the image embedding.
    \item \textbf{Cross-Attention:} Integrates contextual guidance from the condition embedding \(y\).
    \item \textbf{Feed-Forward Networks (FFN):} Enhances feature representations. The FFN hidden layer size is determined by \texttt{mlp\_ratio} \(\times\) \texttt{hidden\_size}. GELU with an approximate tanh calculation is used as the activation function.
    \item \textbf{DropPath Regularization:}  Applies DropPath regularization to prevent overfitting.
\end{itemize}

\vspace{-2mm}
\subsubsection{Linear and Reshape}
After the DiT blocks, a linear layer is applied to the \(S \times D\) output, transforming it to a tensor of shape \(S \times 8C'\). This linear transformation projects each patch embedding from dimension \(D\) to a new dimension of \(8C'\) (equivalent to \(2C' \times P^2\)). Subsequently, this tensor is reshaped to \(2C' \times H' \times W'\), aligning with the VAE's latent spatial dimensions (\(H' \times W'\)) but with twice the channel depth (\(2C'\) instead of \(C'\)).  Within this \(2C' \times H' \times W'\) tensor, the first \(C'\) channels represent the predicted noise, and the subsequent \(C'\) channels represent the predicted variance, primarily used during training.
\subsubsection{VAE Decoder}

At inference time, after this denoising process, the model's output \(2C'\) (only the first \(C'\) channels) is fed into a pre-trained VAE decoder. This decoder then reconstructs an image from this representation, producing a 3-channel image of size \(C \times H \times W\). Finally, this reconstructed image is converted into the final segmentation mask through binarization.

\vspace{-2mm}
\section{Experimental Analysis}
\label{sec:experiments}

\vspace{-4mm}
\subsection{Datasets}
\label{subsec:dataset}
\vspace{-4mm}

The ISIC 2016 \cite{gutman2016skin}, 2017 \cite{codella2018skin}, and 2018 \cite{codella2019skin} challenges are widely recognized benchmarks in medical image segmentation, specifically for skin lesion analysis. These challenges provided large, labeled datasets for detecting and segmenting melanoma from dermoscopic images. The ISIC 2016 dataset included 900 training images and 335 testing images, each with corresponding segmentation masks, allowing evaluation of lesion detection and boundary accuracy. The ISIC 2017 dataset expanded on this with 2000 training images, 150 validation images, and 600 testing images, all with segmentation masks. The ISIC 2018 dataset further increased in size, containing 2594 training images, 100 validation images, and 1000 testing images, ensuring a more robust assessment of segmentation models.

\vspace{-4mm}
\subsection{Metrics}
\label{subsec:metrics}
\vspace{-4mm}

The metrics widely used for evaluating skin lesion segmentation on these datasets include: Dice Similarity Coefficient (Dice), Intersection over Union (IoU), Pixel Accuracy (ACC), Sensitivity (SE), and Specificity (SP). These metrics assess the overlap and classification accuracy between predicted segmentation masks (SM) and ground truth masks (GM).

\begin{align}
\text{Dice} &= \frac{2 |SM \cap GM|}{|SM| + |GM|} \\
\text{IoU} &= \frac{|SM \cap GM|}{|SM \cup GM|} \\
\text{ACC} &= \frac{TP+TN}{TP+FP+TN+FN}  \\
\text{SE} &= \frac{TP}{TP + FN} = \frac{|SM \cap GM|}{|GM|}
\end{align}

\begin{align}
\text{SP} &= \frac{TN}{TN + FP}
\end{align}

Where \(|SM \cap GM|\) is the number of shared pixels, \(|SM|\) and \(|GM|\) are the total pixels in each mask, and TP, TN, FP, and FN are true positives, true negatives, false positives, and false negatives, respectively.

\vspace{-2mm}
\subsection{Implementation Details}
\label{subsec:implementation_details}
\vspace{-2mm}

SegDT was trained with Adam Optimizer with a learning rate of 1e-4 and a batch size of 32 for 100 epochs. The learning rate was reduced by a factor of 10 after 50 epochs. During training, no explicit data enhancement techniques were used. The experiments were carried out on two NVIDIA RTX 3090 GPUs. The input images were bilinearly resized to achieve the target dimensions of the \(256 \times 256\) pixels and then normalized to the [0, 1] range. In the model, these dimensions correspond to \(C=3\), \(H=256\), and \(W=256\). The final segmentation mask was obtained by binarizing the reconstructed image using a fixed threshold of 0.2. This threshold was selected based on performance on a held-out validation set. We explored thresholds ranging from 0.1 to 0.5 in increments of 0.05 and selected 0.2 because it yielded the best overall performance in terms of the Dice score on the validation set. The \texttt{mlp\_ratio} and \texttt{hidden\_size} parameters for the DiT blocks were set to 4 and 192, respectively. 

\vspace{-2mm}
\subsection{Results and Discussion}
\label{subsec:implementation_details}
\vspace{-2mm}

To validate SegDT's accuracy and generalization, we conducted extensive experiments on the ISIC 2016, 2017, and 2018 skin lesion image datasets. We compared SegDT's performance with several state-of-the-art methods and analyzed the performance of different methods under different scenarios and challenging settings.

\vspace{-2mm}
\subsubsection{Segmentation Performance}

Table \ref{tab:combined_isic_results} presents the results of the segmentation in the three ISIC datasets. SegDT achieves state-of-the-art or highly competitive performance across all datasets and most metrics. In ISIC 2016, SegDT achieves the highest Dice score (94. 76\%), IoU (91. 40\%) and accuracy (97. 08\%).  Although DU-Net + achieves slightly higher sensitivity, SegDT demonstrates significantly higher specificity (99. 44\%), indicating its ability to accurately identify healthy tissue, which is crucial in clinical settings.

In ISIC 2017, SegDT again achieved competitive results, with the highest Dice score (91. 70\%) and accuracy (95. 49\%).  Although DU-Net+ shows slightly better IoU and Sensitivity, SegDT maintains significantly higher Specificity (98.74\%).

In the largest dataset, ISIC 2018, SegDT achieves the highest Dice score (94. 51\%) and IoU (90. 43\%), and competitive accuracy and sensitivity.  In particular, SegDT achieves the highest Specificity (97.43\%), further highlighting its strength in identifying healthy tissue.

\vspace{-4mm}
\begin{table}[ht]
\centering
\caption{Segmentation results on the ISIC 2016, 2017, and 2018 datasets.}
\label{tab:combined_isic_results}
\begin{tabular}{lcccccccc}
\toprule
Method & Dataset & Dice$\uparrow$ & IoU$\uparrow$ & ACC$\uparrow$ & SE$\uparrow$ & SP$\uparrow$ & Flops (G)$\downarrow$ & Params (M)$\downarrow$ \\
\midrule
MobileUNETR \cite{perera2024mobileunetr} & 2016 & 92.80 & 87.47 & 96.59 & 93.03 & 96.87 & \textbf{1.30} & \textbf{3.00} \\
GU-Net \cite{cheng2024gu} & 2016 & 89.75 & 82.41 & - & - & - \\
AM-Net \cite{yang2025net} & 2016 & 93.29 & 88.44 & 95.57 & 88.73 & 97.02 & - & - \\
DU-Net+ \cite{kaur2025net+} & 2016 & 92.46 & 85.25 & 96.89 & \textbf{95.75} & 95.64 & 54.00 & 39.00 \\
\textbf{SegDT} & 2016 & \textbf{94.76} & \textbf{91.40} & \textbf{97.08} & 93.35 & \textbf{99.44} & 3.68 & 9.95 \\
\midrule
MobileUNETR \cite{perera2024mobileunetr} & 2017 & 86.84 & 79.00 & 94.46 & 85.18 & 96.93 & \textbf{1.30} & \textbf{3.00} \\
GU-Net \cite{cheng2024gu} & 2017 & 93.94 & 88.98 & - & - & - & - & - \\
PCCTrans \cite{feng2024parallelly} & 2017 & 84.65 & - & 93.28 & - & - & 38.50 & 50.80 \\
AM-Net \cite{yang2025net} & 2017 & 87.30 & 80.02 & 95.13 & 81.70 & 97.99 & - & - \\
DU-Net+ \cite{kaur2025net+} & 2017 & 90.46 & \textbf{85.20} & 95.54 & \textbf{94.07} & 93.85 & 54.00 & 39.00 \\
\textbf{SegDT} & 2017 & \textbf{91.70} & 84.70 & \textbf{95.49} & 87.39 & \textbf{98.74} & 3.68 & 9.95 \\
\midrule
SynergyNet-8s2h \cite{gorade2024synergynet} & 2018 & 89.31 & 80.68 & 94.91 & 87.28 & 97.37 & - & - \\
PCCTrans \cite{feng2024parallelly} & 2018 & 90.03 & - & 96.49 & - & - & 38.50 & 50.80 \\
BRAU-Net++ \cite{lan2024brau} & 2018 & 90.10 & 84.01 & 95.61 & 92.24 & 91.18 & 22.45 & 50.76 \\
MobileUNETR \cite{perera2024mobileunetr} & 2018 & 90.74 & 84.56 & 94.40 & 92.55 & 95.03 & 1.30 & 3.00 \\
GU-Net \cite{cheng2024gu} & 2018 & 92.42 & 86.46 & - & - & - & - & - \\
DU-Net+ \cite{kaur2025net+} & 2018 & 92.93 & 88.13 & \textbf{97.41} & 95.12 & 96.76 & 54.00 & 39.00 \\
SLP-Net \cite{yang2025slp} & 2018 & 88.21 & 80.61 & 93.87 & 89.30 & 95.36 & \textbf{2.30} & \textbf{0.20} \\
\textbf{SegDT} & 2018 & \textbf{94.51} & \textbf{90.43} & 96.81 & \textbf{95.21} & \textbf{97.43} & 3.68 & 9.95 \\
\bottomrule
\end{tabular}
\end{table}

The inclusion of GFLOPs and parameter counts in Table \ref{tab:combined_isic_results} highlights the efficiency of SegDT.  Compared to DU-Net+, SegDT achieves significantly better performance with drastically fewer GFLOPs (3.68 vs. 54.00) and parameters (9.95M vs. 39.00M). This demonstrates SegDT's ability to achieve superior results with a much lighter computational footprint, making it more suitable for resource-constrained environments. Although MobileUNETR and SLP-Net have lower GFLOPs and parameter counts, SegDT offers a better balance between performance and efficiency.  SegDT's performance is achieved with a relatively small increase in computational cost compared to the most efficient models, but with a significant improvement in segmentation accuracy, especially in terms of specificity.  This trade-off is often desirable in medical image analysis, where accuracy is paramount.

SegDT offers a compelling advantage in inference speed compared to IDDPM. SegDT achieves segmentation quality comparable to IDDPM with only 15 inference steps, while IDDPM requires 35 steps. This substantial reduction in steps, facilitated by SegDT's use of rectified flow, reduces the inference time by almost 2 times, making SegDT significantly faster and crucial for practical applications.

\vspace{-4mm}

\subsubsection{Qualitative Analysis}

\begin{figure}
    \centering
    \includegraphics[width=1.0\linewidth]{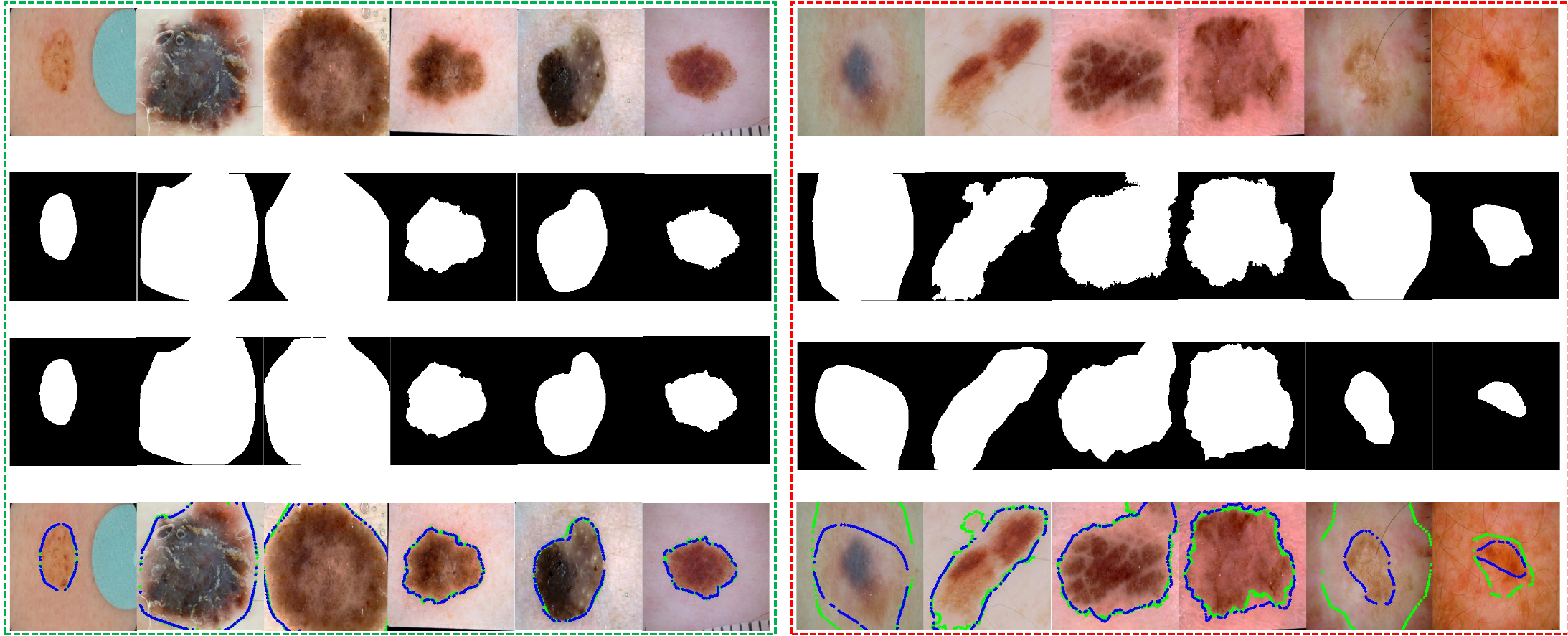}
    \caption{Segmentation results of SegDT on the ISIC datasets. The figure shows examples of well-segmented cases (left) and challenging cases (right).  Each example consists of six columns: the first two are from ISIC 2016, the middle two from ISIC 2017, and the last two from ISIC 2018.  Within each column, the first row shows the original image, the second row the ground truth segmentation mask, and the third row the predicted segmentation mask. The fourth row visualizes the segmentation on the original image, with the ground truth contour in green and the predicted mask contour in blue.}\label{fig:isic_results}
\end{figure}

Figure \ref{fig:isic_results} shows example segmentation results in the ISIC datasets, including both well-segmented cases and challenging cases.  The left examples demonstrate SegDT's ability to accurately delineate lesion boundaries even with variations in shape, size, and texture.  The challenging cases on the right highlight some limitations. For instance, in some cases, SegDT struggles with lesions that have irregular borders or are very small.  This could be due to the limited receptive field of the transformer blocks or the difficulty in capturing fine-grained details in extremely small lesions.  Further investigation is needed to address these limitations.

\vspace{-10mm}
\section{Conclusion}
\label{conclusion}
\vspace{-2mm}

This paper presented SegDT, a novel DiT model for efficient medical image segmentation, specifically targeting skin lesion segmentation.  SegDT leverages a compact DiT architecture and incorporates rectified flow for accelerated inference on resource-constrained GPUs.  Our approach addresses the limitations of existing diffusion models, which often suffer from high computational costs and long inference times, hindering their practical application in real-world clinical settings.

SegDT demonstrates its effectiveness on ISIC datasets (2016, 2017, 2018) by achieving competitive, state-of-the-art segmentation performance with significantly faster inference speeds. This efficiency, enhanced by the use of rectified flow, is crucial for rapid clinical analysis in real-world applications. Furthermore, SegDT's compact architecture enables deployment on low-cost GPUs, broadening the accessibility of advanced medical image segmentation and contributing to the democratization of cutting-edge diagnostic tools.

Future work will focus on further optimizing SegDT's architecture and exploring additional techniques to enhance its performance and efficiency.  We also plan to investigate the model's generalizability to other medical image segmentation tasks and datasets.  Furthermore, we will explore incorporating additional clinical information, such as patient metadata, to further improve segmentation accuracy and clinical utility.  We believe that SegDT represents a promising step towards the development of efficient and accurate medical image segmentation models for real-world clinical applications.

\vspace{-4mm}
\bibliography{references}
\bibliographystyle{splncs04}

\end{document}